\documentclass{article}

\usepackage{PRIMEarxiv}

\usepackage[utf8]{inputenc} 
\usepackage[T1]{fontenc}    
\usepackage{hyperref}       
\usepackage{url}            
\usepackage{booktabs}       
\usepackage{amsmath, amsfonts}       
\newtheorem{thm}{Theorem}
\usepackage{amssymb}
\usepackage{nccmath}
\usepackage{subfigure}
\usepackage{algorithmic}
\usepackage{algorithm}
\usepackage{nicefrac}       
\usepackage{microtype}      
\usepackage{lipsum}
\usepackage{fancyhdr}       
\usepackage{graphicx}       
\graphicspath{{media/}}     

\DeclareMathOperator*{\argmina}{\arg\min}
\title{Neural Message Passing for Objective-Based Uncertainty Quantification and Optimal Experimental Design
}

\author{
  Qihua Chen \\
  University of Science and Technology of China \\
  Hefei, Anhui, CN 230027\\
  \texttt{cqh@mail.ustc.edu.cn} \\
   \And
  Xuejin Chen \\
  University of Science and technology of China \\
  Hefei, Anhui, CN 230027\\
  \texttt{xjchen99@ustc.edu.cn} \\
   \And
  Hyun-Myung Woo\thanks{Corresponding authors}\\
  Incheon National University\\
  Incheon 22012, South Korea\\
  \texttt{hmwoo@inu.ac.kr} \\
    \And
  Byung-Jun Yoon\footnotemark[1] \\
  Brookhaven National Laboratory\\
  Upton, NY 11973, USA\\
  Texas A\&M University \\
  College Station, TX 77843, USA\\
  \texttt{bjyoon@ece.tamu.edu} \\
}

\begin{document}
\maketitle

\newcommand{\tcr}{}
\newcommand{\tcrtwo}{}
\begin{abstract}
Various real-world scientific applications involve the mathematical modeling of complex uncertain systems with numerous unknown parameters. Accurate parameter estimation is often practically infeasible in such systems, as the available training data may be insufficient and the cost of acquiring additional data may be high. In such cases, based on a Bayesian paradigm, we can design robust operators retaining the best overall performance across all possible models and design optimal experiments that can effectively reduce uncertainty to enhance the performance of such operators maximally. \tcrtwo{While objective-based uncertainty quantification (objective-UQ) based on MOCU (mean objective cost of uncertainty) provides an effective means for quantifying uncertainty in complex systems, the high computational cost of estimating MOCU has been a challenge in applying it to real-world scientific/engineering problems.} In this work, we propose a novel scheme to reduce the computational cost for objective-UQ via MOCU based on a data-driven approach. We adopt a neural message-passing model for surrogate modeling, incorporating a novel axiomatic constraint loss that penalizes an increase in the estimated system uncertainty. As an illustrative example, we consider the optimal experimental design (OED) problem for uncertain Kuramoto models, where the goal is to predict the experiments that can most effectively enhance robust synchronization performance through uncertainty reduction. We show that our proposed approach can accelerate MOCU-based OED by four to five orders of magnitude, without any visible performance loss compared to the state-of-the-art. The proposed approach applies to general OED tasks, beyond the Kuramoto model.
\end{abstract}


\section{Introduction}
Real-world scientific and engineering applications often involve modeling complex uncertain systems with substantial uncertainties. \tcr{For such systems, the amount of data available for learning the corresponding models may pale when compared to the model complexity, which may render accurate estimation of the model parameters practically infeasible}. In the presence of significant model uncertainty, designing operators--such as predictors or controllers--based on the best (\tcrtwo{i.e.}, most likely) model may be unreliable. This issue may be alleviated by taking a Bayesian approach to consider model uncertainty to design a robust operator that can guarantee good average performance despite the uncertainty. \tcr{Generally, robust operators have the advantage of preventing catastrophic failures caused by model mismatches and maintaining acceptable performances under model uncertainty. However, the overall performance will take a hit--at least to some extent, even if the expected degradation is acceptable--when compared to the optimal operator, whose design requires full knowledge of the true underlying model.} 

\tcr{In this context, optimal experimental design (OED) aims to design, prioritize, or select experiments whose outcome is expected to effectively reduce the uncertainty in the current model such that the performance of the robust operator designed thereupon can be optimally improved.
In general, an OED method first quantifies the uncertainty of the current model. Then, the expected efficacy of each potential experiment in the experimental design space is predicted, which can be used to intelligently select the best experiment that is expected to maximally reduce the model uncertainty that critically impacts the performance of the robust operator. Once the optimal experiment is designed (or selected), one may conduct the experiment at hand and use the actual experimental outcome to reduce model uncertainty. Within the available experimental budget, this OED cycle--which consists of \textit{designing an experiment}, \textit{conducting the designed experiment}, and \textit{updating the model based on the experimental outcome} (and thereby reducing model uncertainty)--may be repeated.
It should be noted that the experiment designed by an OED scheme may not necessarily be optimal, as the optimality of the experiment depends on the true model, which is \textit{unknown}. However, \textit{on average}, experiments designed via OED can reduce model uncertainty much more rapidly compared to experiments that are either randomly selected or designed in a heuristic manner. Consequently, OED can minimize the experimental cost and time required to reduce the model uncertainty--or equivalently, improve the performance of the robust operator designed based on an uncertain model--to the desired level.}

\tcr{In the context of OED, one fundamental question emerges naturally: how can one quantify the model uncertainty? In Bayesian OED, the model uncertainty has been quantified via variance of the posterior distribution of the parameters. To predict the potential efficacy of each experiment, the so-called \textit{expected information gain (EIG)} has been widely used~\cite{ryan2003estimating, long2013fast, tsilifis2017efficient, beck2018fast} and has been shown to be effective, especially, when the objective is better identification or more accurate modeling of the system of interest. However, in real-world scientific applications where one may want to learn models for specific tasks or operations--\tcrtwo{e.g.}, control or classification--the aforementioned approaches may not always be effective as the reduced uncertainty may not directly help perform the tasks at hand or achieve the operational goals.}

\tcr{The mean objective cost of uncertainty (MOCU)~\cite{Yoon2013,Yoon2021} addresses this shortcoming by quantifying the expected increase in the operational cost that is induced by model uncertainty, providing an effective means for objective-based uncertainty quantification (objective-UQ).} As MOCU captures the model uncertainty that impacts the performance of the operator designed based on the model, it leads to an effective framework for optimal experimental design (OED)~\cite{Hong2021,Dehghannasiri2014tcbb}, based on which the expected efficacy of potential experiments may be quantified. \tcr{To date, MOCU-based OED has been widely applied to a number of applications~\cite{Dehghannasiri2014tcbb,Broumand2015,Dehghannasiri2015bmc,Talapatra2018,Zhao2020TSP,Zhao2021ICLR}~\cite{Zhao2021AISTATS,Zhao2021NeurIPS}. For example, MOCU-based OED was applied to pool-based Bayesian active-learning problems, where the objective is to improve the sample labeling efficiency~\cite{Zhao2021ICLR,Zhao2021AISTATS,Zhao2021NeurIPS}. MOCU-based acquisition functions--such as weighted MOCU~\cite{Zhao2021ICLR} or soft MOCU~\cite{Zhao2021AISTATS,Zhao2021NeurIPS}--were defined, which were then used to identify and label the optimal sample expected to maximally reduce the classification error. In this setup, an experiment corresponds to selecting an unlabeled data point from a given pool and the experimental outcome is the true label of the selected data point.
Furthermore, MOCU-based UQ and OED were also used to investigate the problem of designing a robust filter that can reliably predict the drug concentration level via uncertain stochastic differential equation (SDE)-based pharmacokinetic models~\cite{Zhao2020TSP}. In the study, the experimental design concerned selecting the best parameter to query for further measurement. The outcome of the optimal experiment--\tcrtwo{i.e.}, the additional measurement data of the selected parameter--is aimed at maximally reducing the prediction error of the robust filter.}

\tcrtwo{Recently,} the MOCU-based OED strategy has been developed for uncertain Kuramoto models~\cite{Hong2021}, which consist of interconnected oscillators described by coupled ordinary differential equations (ODEs). Kuramoto models~\cite{Kuramoto1975} have been extensively studied in various fields, thanks to their ability to simulate interesting collective behavior in complex networked systems such as the robustness of microgrid networks~\cite{Skardal2015} or synchrony in brain networks~\cite{Lehnertz2009,Mohseni2017}. Given an uncertain Kuramoto model, the objective of the experimental design was to predict the optimal experiment that can most effectively reduce the control cost to achieve robust synchronization of the uncertain Kuramoto oscillators.

\tcrtwo{One practical challenge of MOCU-based OED is the high computational complexity that mainly originates from the complexity of predicting the expected efficacy ({i.e.}, the expected remaining MOCU) of each experiment in the experimental space. In general, real-world scientific/engineering applications are designed based on highly complex operations, resulting in no closed-form expression for (expected remaining) MOCU. For example, in the original OED study for the robust control of uncertain Kuramoto models~\cite{Hong2021}, for a given experiment, the expected remaining MOCU was estimated via Monte Carlo sampling~\cite{metropolis1949monte}, where each sample defining a Kuramoto model involves repetitively solving coupled differential equations to compute the operational cost of the best operator that is optimal for the given Kuramoto model. In order to reduce the effective time complexity, parallelism was considered by computing the optimal operational cost of each Kuramoto model on different processing units. However, even with parallelism, MOCU estimation remained highly complex, limiting the applicability and scalability of the MOCU-based OED. In order to alleviate the computational bottleneck, a machine learning~(ML)-based approach was proposed in~\cite{Woo2021}. In this work, a fully connected neural network model was trained to predict--without solving the ODE--whether or not a given Kuramoto model reaches frequency synchronization. This approach remarkably reduced the cost of predicting the optimal operator, thereby significantly accelerating the MOCU computation and expediting the MOCU-based OED by two orders of magnitude without any degradation in OED performance. However, the approach in~\cite{Woo2021} still resorted to Monte Carlo sampling for MOCU estimation, and it remained an open question whether it is possible to design accurate deep surrogate models for direct MOCU estimation by building on recent advances in deep learning (DL) techniques~\cite{Gilmer2017, Federated, Federated2}. Such deep surrogate models would obviate the need for costly Monte Carlo sampling and further accelerate MOCU-based OED.}

\tcrtwo{In this work, we address this open problem by proposing the use of a deep learning technique to build an effective surrogate model for MOCU computation for the first time. We show that our proposed scheme is capable of accelerating MOCU-based OED by four to five orders of magnitude by eliminating the Monte Carlo sampling process and the need for repetitively solving ordinary differential equations for the uncertain Kuramoto models.
Our main contributions are as follows. First, we propose a novel sampling-free approach that can effectively address the computational bottleneck of MOCU estimation by directly estimating MOCU for a given uncertainty class via neural message-passing~\cite{Gilmer2017}. Second, we introduce a novel axiomatic constraint loss that improves the estimation accuracy of the message-passing model to enhance the experimental design. Experiments show that our proposed approach accelerates OED up to five orders of magnitude while maintaining virtually identical performance compared to the original sampling-based approach~\cite{Hong2021}.} To the best of our knowledge, this is the first work that demonstrates the possibility of designing data-driven surrogate models for efficient experimental design based on objective-UQ via MOCU, \tcrtwo{thereby laying the foundation for a wide range of real-world applications of MOCU-based OED.}

\tcr{The current paper is organized as follows. In Section~\ref{sec:oed}, we provide the details of MOCU-based OED for uncertain Kuramoto models, where the operational goal is to design an effective control strategy with minimum cost to achieve robust synchronization\tcrtwo{, which was originally formulated in~\cite{Hong2021}. We also analyze the time complexity of MOCU estimation using the sampling-based method in~\cite{Hong2021}}. In Section~\ref{sec:UQ_MPNN}, we propose a DL approach for quantifying model uncertainty via MOCU. Specifically, we build a DL surrogate model using a message-passing neural network (MPNN) that directly quantifies MOCU, which then can be used for designing effective experiments that can enhance the performance and reduce the cost of control for uncertain Kuramoto models. We demonstrate the efficacy of the proposed approach in Section~\ref{sec:simulation} based on comprehensive simulation results. \tcrtwo{We conclude the paper in Section~\ref{sec:conclusion} with important insights learned from this study, discussion of the limitations of the current work, and future research directions}.}

\section{Optimal Experimental Design for Uncertain Kuramoto model}\label{sec:oed}
\tcr{In this section, we present a brief review of the MOCU-based OED scheme for the robust synchronization of uncertain Kuramoto models, originally considered in~\cite{Hong2021}. First, we describe the uncertainty class of Kuramoto models. Based on this, we outline the OED setup, where the objective is the robust synchronization of an uncertain Kuramoto model. Next, we describe how one can prioritize the potential experiments based on the (expected remaining) MOCU. We then present how MOCU can be estimated via sampling, as proposed in a prior study~\cite{Hong2021} and analyze the computational complexity of the sampling-based MOCU estimation and experimental design. As the primary goal of the current study is to accelerate MOCU-based UQ and OED by leveraging a DL model, the latest sampling-based approach in~\cite{Hong2021} serves as the baseline to which we compare our proposed approach.}

\subsection{\tcr{Robust synchronization of uncertain Kuramoto model}}\label{UK}\vspace{5pt}
\tcr{The Kuramoto model consists of ${N}$ interacting oscillators described by ${N}$ coupled ordinary differential equations:}
\begin{equation}
\dot{\theta_{i}}(t) = \omega _{i} + \sum_{j=1}^{N} a_{i,j} \sin(\theta_{j}(t) - \theta_{i}(t)), \quad\tcr{i = 1, 2, \dots, N,}\label{eq:kuramoto}
\end{equation}
where $N$ is the number of oscillators, $\theta_{i}(t)$ is the instantaneous phase of the ${i}$th oscillator with the intrinsic natural frequency $\omega_{i}$, and $a_{i,j} $ $(=a_{j,i})$ represents the coupling strength between the ${i}$th and ${j}$th oscillators.
The model has been widely used to describe and investigate the collective behavior of constituents of a complex networked system, where examples include brain networks~\cite{Hammond2007,Breakspear2010,Schmidt2015}, power systems~\cite{Guo2021}, microgrids~\cite{Skardal2015}, and synchronously flashing fireflies~\cite{Buck76}. Of special interest has been the synchronization phenomena of Kuramoto models, which have been shown to be related to emergent system-level properties--such as the stability of microgrid networks~\cite{Skardal2015} or neurological disorders in brain networks~\cite{Lehnertz2009,Mohseni2017}. 

\tcr{In a recent study~\cite{Hong2021}, the optimal experimental design problem for uncertain Kuramoto models was investigated. The task of interest was to achieve global synchronization of a given Kuramoto model through external control in the presence of model uncertainty, where MOCU-based OED was used to select experiments that can effectively reduce the uncertainty that affects this goal. In our current study, we also consider this robust global synchronization problem, in order to demonstrate how a DL-based approach can significantly improve the computational efficiency of OED without degrading the overall efficacy.}

\tcr{As in~\cite{Hong2021}, we assume that the coupling strength \tcr{$a_{i,j}, i, j = 1, 2, \dots, N,$} is uncertain and known only up to a range $a_{i,j}\in {[ a_{i,j}^{\ell} a_{i,j}^u ] }$. This gives rise to an uncertainty class $\mathcal{A}$, which consists of all possible parameter vector, ${\mathbf{a} = [a_{1, 2}, a_{1, 3}, \dots, a_{N-1, N}]^T} \in {\mathbf{\mathcal{A}}}$, distributed according to a prior distribution ${P_{\mathbf{\mathcal{A}}} ( \mathbf{a} )}$ consistent with the given constraints.
Given a Kuramoto model with uncertain coupling strength $\mathbf{a}\in{\mathbf{\mathcal{A}}}$, our operational objective is to apply robust control to ensure that the oscillators are frequency synchronized such that}
\begin{equation} 
\lim_{t \to \infty} | \dot{\theta}_i \left(t\right) - \dot{\theta}_j \left(t\right) | = 0, \quad \forall i,j.
\end{equation}
Specifically, we consider the control scheme in~\cite{Hong2021}, where a control oscillator with natural frequency $\omega _{N+1}$ is added to the uncertain Kuramoto model to interact with the ${N}$ oscillators with uniform interaction strength $a_{N+1}$ (i.e., ${a_{N+1} = a_{N+1, i}, \forall i}$). The extended model with the control oscillator is given by:
\begin{equation} \label{control}
\begin{split}
\dot \theta_i&(t) =  \omega_i + \sum_{j=1}^{N} a_{i,j} \sin (\theta_j(t)- \theta_i(t))+ a_{N+1} \sin (\theta_{N+1}(t)- \theta_i(t)), \quad i=1,2,\dots,N+1.
\end{split}
\end{equation}
\tcr{For a Kuramoto model with a given interaction strength $\mathbf{a}$, we define $\xi(\mathbf{a})$ as the minimum (non-negative) interaction strength that ensures the global synchronization of the extended Kuramoto model when the interaction strength $a_N$ of the control oscillator is set to $a_{N+1}=\xi(\mathbf{a})$}. We also regard $\xi(\mathbf{a})$ as the cost of control since exerting stronger control would require higher power in practice.

In the presence of uncertainty $\mathcal{A}$, the cost of \tcr{\textit{robust}} control that guarantees model synchronization would inevitably increase as $a_{N+1}$ needs to be large enough to ensure synchronization for all possible models in the uncertainty class. For this reason, the optimal robust control (and the cost thereof) is defined as follows~\cite{Hong2021}:
\begin{equation}
\xi^*(\mathcal{A})=\max_{\mathbf{a}\in \mathcal{A}}\xi(\mathbf{a}).
\end{equation}
MOCU~\cite{Yoon2013,Yoon2021} captures the expected differential cost induced by model uncertainty due to having to use the optimal \textit{robust} control (that maintains good average control performance across the uncertainty class $\mathcal{A}$) instead of the optimal control that is model-specific (which cannot be used in practice as the true model is unknown) as follows:
\begin{equation}\label{MOCU}
    \mathnormal{M}(\mathcal{A}) = \mathop{\mathbb{E}}_{\mathbf{a}\in \mathcal{A}} [\xi^*(\mathcal{A}) - \xi(\mathbf{a})].
\end{equation}

\begin{figure*}
    \centering
    \includegraphics[width=0.75\textwidth]{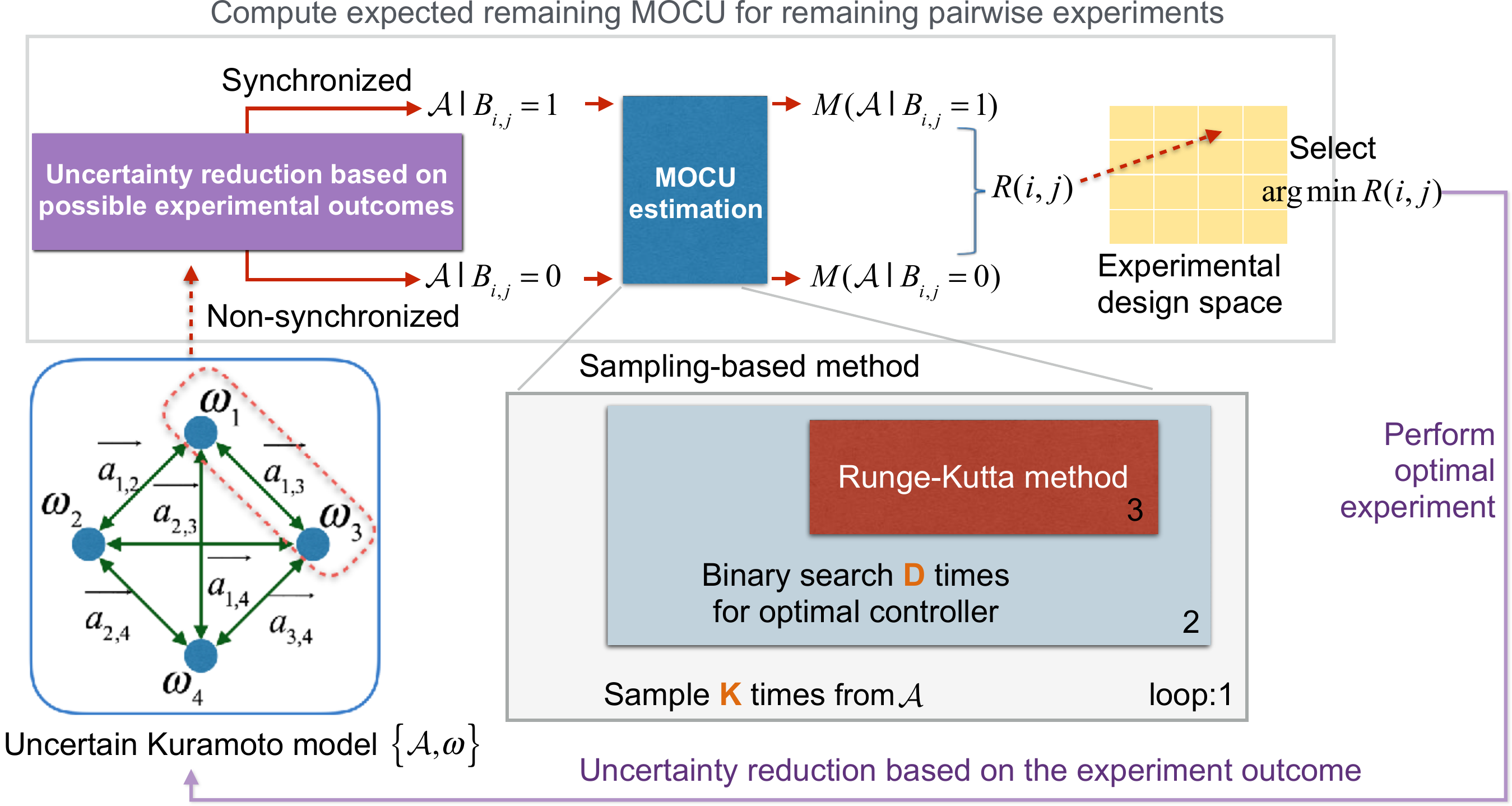}\vspace{10pt}
    \caption{Overview of the MOCU-based optimal experimental design for robust synchronization of uncertain Kuramoto models, where MOCU is estimated based on the sampling-based approach~\cite{Hong2021}.}
    \label{OEDpipe}
\end{figure*}

\subsection{Experimental design space} \label{OED}\vspace{5pt}
\tcr{In this study, we consider the same experimental design space as in~\cite{Hong2021}. Each experiment in this design space consists of the following steps. First, we select a pair of oscillators ${\left(i, j\right)}$ in the original Kuramoto model that consists of ${N}$ oscillators and isolate the pair from the ${N-2}$ remaining oscillators. Next, we observe whether or not the isolated oscillator pair ${\left(i, j\right)}$ gets spontaneously synchronized without external control. According to the binary experimental outcome (i.e., whether the oscillator pair gets synchronized or not), we can reduce the model uncertainty by utilizing Theorem~\ref{theo:sync} in~\cite{Hong2021} (shown below as a reference for readers):}
\begin{thm} \label{theo:sync}
Assume the Kuramoto model of two-oscillators:
\begin{equation}\label{Ku2}
\begin{split}
\dot \theta_1 \left(t\right)&= \omega_1 + a \sin\left(\theta_2\left(t\right)- \theta_1\left(t\right)\right),\\
\dot \theta_2 \left(t\right)&= \omega_2 + a \sin\left(\theta_1\left(t\right) - \theta_2\left(t\right)\right),
\end{split}
\end{equation}
with initial phases $\theta_1\left(0\right), \theta_2\left(0\right)\in[0,2\pi)$. For any solutions $\theta_1 \left(t\right)$ and $\theta_2\left(t\right)$ to~\eqref{Ku2}, there holds 
$|\dot\theta_1(t) - \dot\theta_2(t) | \to 0$ as $t\to\infty$ 
if and only if 
$(1/2)*|\omega_1-\omega_2|\le a$. \hspace{1.22in} $\hfill\blacksquare$
\end{thm}\vspace{5pt}

Based on Theorem~\ref{theo:sync}~\cite{Hong2021}, the model uncertainty can be reduced by increasing the lower bound of the uncertain interval ${a^\ell_{i,j}}$ to ${\max((1/2)*|\omega_i - \omega_j|,a^\ell_{i,j})}$ if the oscillator pair gets synchronized. Otherwise, the upper bound ${a^u_{i,j}}$ is reduced to ${\min((1/2)*|\omega_i - \omega_j|,a^u_{i,j})}$.\vspace{5pt}

\tcr{\subsection{MOCU-based optimal experimental design} \label{MOCU-basedOED}}
\vspace{3pt}
\tcr{Figure~\ref{OEDpipe} provides an overview of the MOCU-based OED scheme, where MOCU is computed using a Monte Carlo estimator~\cite{Hong2021}. The MOCU-based OED provides a systematic means to identify the optimal experiment that is expected to most sharply reduce the uncertainty affecting the cost of the operation of interest--namely, the cost of achieving robust synchronization through control. To prioritize experiments, we first compute the \textit{expected remaining MOCU} for every pairwise synchronization experiment, which is given by:}

\begin{equation}
\begin{split}
    R(i,j) &= \mathop{\mathbb{E}}_{B_{i,j}}[M(\mathcal{A}|B_{i,j})] = \sum_{b \in {0,1}} P(B_{i,j}=b)M(\mathcal{A}|B_{i,j}=b),
\end{split}
\end{equation}
where $\mathcal{A}|B_{i,j}$ is the remaining uncertainty conditioned on the experimental outcome $B_{i,j}$. The probability $P\left(B_{i, j} = b\right)$ is given by:
\begin{equation}
    \begin{aligned}
        P\left(B_{i, j} = 1\right) = \frac{a^u_{i,j} - \tilde{a}_{i,j}}{a^u_{i,j} - a^\ell_{i,j}}, \quad
        P\left(B_{i, j} = 0\right) = \frac{\tilde{a}_{i,j}-a^\ell_{i,j} }{a^u_{i,j} - a^\ell_{i,j}},
    \end{aligned}
\end{equation}
where ${\tilde{a}_{i,j} = \min \left( \max\left( (1/2)*|\omega_i - \omega_j|, a^\ell_{i,j} \right) , a^u_{i, j}\right)}$.
Here we assume that the uncertain interaction strength $a_{i,j}$ is uniformly distributed between the bounds $a_{i,j}^\ell$ and $a_{i,j}^u$ and that the experimental outcome $B_{i,j}$ is used to update either the lower or upper bound according to Theorem~\ref{theo:sync} as described in Sec.~\ref{OED}.
The pairwise synchronization experiment for the oscillator pair ${\left(i, j\right)}$ yielding the minimum expected remaining MOCU $R(i,j)$ is selected as the optimal experiment ${\left(i^*,j^*\right)}$:
\begin{equation}
    \left(i^*,j^*\right) = \argmina_{1\leq i<j\leq N }R(i,j).\label{eq:OEDProblem}
\end{equation}

Ideally, as described in Algorithm~\ref{OED_code}, the expected remaining MOCU should be re-estimated after each experiment, based on the reduced uncertainty class updated according to the outcome of the previous experiment. However, since this increases the overall computational cost of the experimental design procedure, one may take a simpler strategy if desired, where all experiments are prioritized based on their respective expected remaining MOCU values estimated based on the initial uncertainty class. Although this approach is clearly sub-optimal from a theoretical point of view, the strategy has been shown to lead to virtually identical performance compared to the optimal strategy in a number of examples considered in previous studies~\cite{Hong2021,Woo2021}.
\begin{algorithm}[H]
\caption{MOCU-based optimal experimental design (OED) for robust synchronization of uncertain Kuramoto model with ${N}$ oscillators}\label{OED_code}
    \begin{algorithmic}
    \STATE
    \STATE {\textbf{Input:}} Uncertain Kuramoto model \{$N, \mathcal{A}$, $\boldsymbol{\omega}$\}
    \STATE experimental design space $ \mathcal{E} = \{(1,2), ...,(N-1,N)\}$
    \STATE {\textbf{Output:}} Reduced uncertainty class $\mathcal{A}$
    \STATE
    \STATE\hspace{0.0cm} \textbf{while} $\mathcal{E}$ is not empty \textbf{do}
    
        
        \STATE\hspace{0.5cm} /$\ast$ Compute expected remaining MOCU $\ast$/
        \STATE\hspace{0.5cm} \textbf{for} $(i,j)$ in $\mathcal{E}$ \textbf{do}
            \STATE\hspace{1.0cm} $ \tilde{a}_{i,j} \leftarrow \min \left( \max\left( \frac{1}{2} |\omega_i - \omega_j |, a^\ell_{i,j} \right) , a^u_{i, j}\right) $
            \STATE\hspace{1.0cm} $P(B_{i,j}=1) \leftarrow \frac{a^u_{i,j} - \tilde{a}_{i,j}}{a^u_{i,j} - a^\ell_{i,j}}$
            \STATE\hspace{1.0cm} $P(B_{i,j}=0) \leftarrow \frac{\tilde{a}_{i,j}-a^\ell_{i,j} }{a^u_{i,j} - a^\ell_{i,j}}$
            \STATE\hspace{1.0cm}     $R(i,j) \leftarrow \sum_{b \in {0,1}} P(B_{i,j}=b)M(\mathcal{A}|B_{ij}=b)$
        \STATE\hspace{0.5cm} \textbf{end for}
        \STATE\hspace{0.5cm} /$\ast$ Select and conduct the optimal experiment $\ast$/
        \STATE\hspace{0.5cm} $(i^*,j^*)\leftarrow \argmina_{(i,j) \in \mathcal{E} }R(i,j)$\vspace{1pt}
        \STATE\hspace{0.5cm} Conduct the optimal experiment $(i^*,j^*)$
            \STATE\hspace{0.5cm} \textbf{if} $(i^*,j^*)$ gets synchronized \textbf{then}
                \STATE\hspace{1.0cm} $\mathcal{A}\ni a_{i^*,j^*}^{\ell} \leftarrow \tilde{a}_{i^*,j^*}$
            \STATE\hspace{0.5cm} \textbf{else}
                \STATE\hspace{1.0cm} $\mathcal{A}\ni a_{i^*,j^*}^u \leftarrow \tilde{a}_{i^*,j^*}$
            \STATE\hspace{0.5cm} \textbf{end if}
            \STATE\hspace{0.5cm} remove $(i^*,j^*)$ from $\mathcal{E}$
    \STATE\hspace{0.0cm} \textbf{end while}
    \STATE\hspace{0.0cm} \textbf{return} $\mathcal{A}$
    \end{algorithmic}
\end{algorithm}

\subsection{Computational complexity} \label{TC}\vspace{5pt}
One major obstacle in applying MOCU-based OED to real-world applications has been the high computational cost of computing the expected remaining MOCU for the experiments under consideration. In many cases, there may be no closed-form expression of MOCU, and it may have to be numerically estimated as in~\cite{Hong2021}. 
The sampling-based approach~\cite{Hong2021} for MOCU estimation involves repeated sampling from the uncertainty class (to obtain a sample estimate of the expected cost increase) as well as a binary search to find the optimal operational cost.

\begin{algorithm}[t]
\caption{Sampling-based MOCU computation}\label{MOCU_code}
    \begin{algorithmic}
    \STATE
    \STATE {\textbf{Input:}} Uncertain Kuramoto model \{$N, \mathcal{A}$, $\boldsymbol{\omega}$\}, natural frequency of controlling oscillator $\Bar{\omega}$
    \STATE {\textbf{Output:}} ${M\left(\mathcal{A}\right)}$
    \STATE
    \STATE\hspace{0.0cm} \textbf{for} {$i = 1 \to K$}
        \STATE\hspace{0.5cm} draw a sample $\mathbf{a}_i$ from $ P\left(\mathcal{A}\right) $
        \STATE\hspace{0.5cm} /$\ast$ Initialize binary search space $\ast$/
        \STATE\hspace{0.5cm} $a_{N+1}^u \leftarrow 2$\\\vspace{2pt}
        \STATE\hspace{0.5cm} \textbf{while} $\{\mathbf{a}_i, a_{N+1}^u, \boldsymbol{\omega}, \Bar{\omega}\}$ is not synchronized \textbf{do}\vspace{2pt}
        \STATE\hspace{1.0cm} $a_{N+1}^u \leftarrow a_{N+1}^u + 2$\vspace{2pt}
        \STATE\hspace{0.5cm} \textbf{end while}
        \STATE\hspace{0.5cm} $a_{N+1}^{\ell} \leftarrow 0$ \\\vspace{2pt}
        \STATE\hspace{0.5cm} /$\ast$ Binary search for $\xi(a_i)$ $\ast$/
        \STATE\hspace{0.5cm} \textbf{while} {$a_{N+1}^u - a_{N+1}^{\ell} < 2.5e^{-4}$} \textbf{do}
            \STATE\hspace{1.0cm} $c \leftarrow \frac{a_{N+1}^u + a_{N+1}^{\ell}}{2}$\vspace{2pt}
            \STATE\hspace{1.0cm} $state \leftarrow Runge\-Kutta Solver(\mathbf{a}_i, c, \boldsymbol{\omega}, \Bar{\omega})$
            \STATE\hspace{1.0cm} \textbf{if} $state$ is synchronize \textbf{then}
                \STATE\hspace{1.5cm} $a_{N+1}^u \leftarrow c$
            \STATE\hspace{1.0cm} \textbf{else}
                \STATE\hspace{1.5cm} $a_{N+1}^{\ell} \leftarrow c$\vspace{2pt}
            \STATE\hspace{1.0cm} \textbf{end if}
        \STATE\hspace{0.5cm} \textbf{end while}
    \STATE\hspace{0.5cm} $\xi(\mathbf{a}_i) \leftarrow c$
    \STATE\hspace{0.0cm} \textbf{end for}
    \STATE\hspace{0.0cm} \textbf{return} $\sum_{i=1}^K \frac{1}{K}(\mathop{max}\limits_{1\leq j \leq K}\xi(\mathbf{a}_j)-\xi(\mathbf{a}_i))$
    \end{algorithmic}
\end{algorithm}
The sampling-based MOCU estimation process is summarized in Algorithm~\ref{MOCU_code}. Given an uncertainty class $\mathcal{A}$, $K$ sample points are randomly drawn from $P\left(\mathcal{A}\right)$, where each point corresponds to one possible Kuramoto model within the given uncertainty class. Generally, as the sample size $K$ increases, the accuracy of the estimated MOCU improves. Finding the optimal control (and its cost) for each sample point (i.e., a specific Kuramoto model in $\mathcal{A}$) requires numerical optimization. In~\cite{Hong2021}, the optimal control was predicted via binary search. As a consequence, the overall computational complexity for sampling-based MOCU estimation is given by $\mathcal{O}\left(KDMN^2\right)$, where ${D}$ is the average complexity of the binary search, and ${M}$ is the time duration for solving the ordinary differential equations (e.g., a fourth-order Runge-Kutta method was used in~\cite{Hong2021}), and ${N}$ is the number of oscillators in the original Kuramoto model. Furthermore, to estimate the expected remaining MOCU for a given experiment, we need to estimate the MOCU conditioned on every possible experimental outcome (which is binary in this case). Finally, this process needs to be repeated for all $N \choose 2$ pairwise experiments to be able to identify the optimal experiment with the smallest $R(i,j)$.

As described in the previous section, this entire process--estimating the expected remaining MOCU for all (remaining) experiments in the experimental design space, identifying the optimal experiment, conducting the selected experiment, and reducing the uncertainty class according to the experimental outcome--may be repeated until we have performed all possible experiments or until we have exhausted the experimental budget. This will increase the computational complexity of OED even further.

\section{Message-Passing Neural Network for Uncertainty Quantification}\label{sec:UQ_MPNN}

In this section, we propose a data-driven approach for accelerating objective-UQ, thereby significantly expediting MOCU-based OED. At the heart of the proposed method lies a message-passing neural network (MPNN)~\cite{Gilmer2017} that incorporates a novel axiomatic constraint. MPNN is a graph neural network-based regression model that can predict the output by repeatedly exchanging messages between neighboring nodes that are connected in the input undirected weighted graph. Similar architectures have been shown to achieve superior performance over other existing schemes in molecular analysis~\cite{Gilmer2017,duvenaud2015convolutional}, recommender systems~\cite{berg2017graph}, and visual scene understanding~\cite{xu2017scene,zhang2021holistic}.
\tcr{In what follows, we first describe the MPNN architecture used to directly quantify (expected remaining) MOCU for an uncertain Kuramoto model. Next, we propose a novel constraint function that penalizes the deviation from an axiom regarding MOCU. Finally, we present the training procedure of the proposed MPNN model.}

\subsection{Model architecture}\vspace{5pt}

\begin{figure*}[t]
    \centering
    \includegraphics[width=1.0\textwidth]{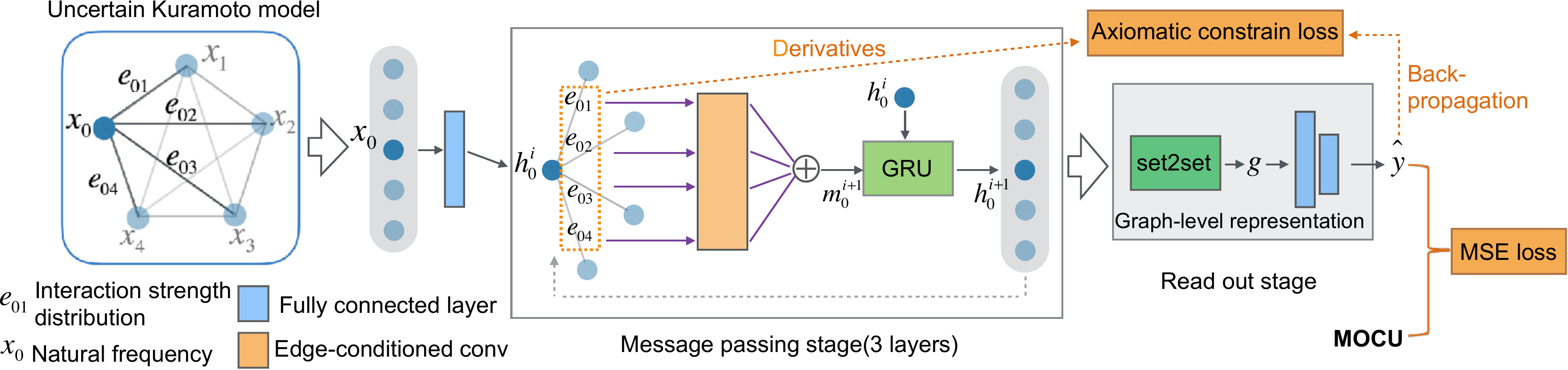}
    \caption{Neural message-passing architecture for uncertainty quantification of the uncertain Kuramoto model.}
    \vspace{-5pt}
    \label{Architecture}
\end{figure*}

\tcr{Figure~\ref{Architecture} illustrates the overall architecture of the MPNN model that takes an uncertain Kuramoto model, represented as ${\{\{\omega_i\}^{N}_{i=1}, \mathcal{A}\}}$, as its \textit{input} and predicts the MOCU for the given uncertain model as the \textit{output}. Specifically, we set the initial node features $x_v$ and edge features $e_{vw}$ of the MPNN model to the natural frequency of the oscillators and the lower and upper bounds of the interaction strength of the uncertain Kuramoto model, respectively.} The model operates in two phases: a \textit{message-passing} phase and a \textit{readout} phase. The message-passing phase recurrently interchanges information across the graph to build a node-level neural representation of the input uncertain Kuramoto model. The readout phase aggregates the node-level representation of the Kuramoto model to a graph-level representation and makes a prediction. The message-passing phase runs for $T$ repeated steps with message function $M_t$ and vertex update function $U_t$, where $T=3$ was used in this study. During the message-passing phase, for each vertex $v$, a message $m_v^t$ is aggregated from the neighbors through the edges, and the vertex hidden states $h_v^t$ is then updated based on $m_v^t$ as follows:
\begin{equation}
    m_v^{t+1} = \sum_{w\in N(v)} M_t(h_v^t, h_w^t, e_{vw})
\end{equation}
\begin{equation}
    h_v^{t+1} = U_t(h_v^t, m_v^{t+1}),
\end{equation}

where $N(v)$ is the set of neighbors of $v$ in the graph. $h_v^0$ is initialized as a learnable linear mapping of the initial node features $x_v$ followed by the ReLU (Rectified Linear Unit) activation. Here, the message function $M_t$ is an Edge-Conditioned Convolution~\cite{Simonovsky2017ecc}, a learnable network that dynamically generates filters based on the edge features. We employ a Gated Recurrent Unit (GRU)~\cite{GRU} as the update function $U_t$ which applies gating mechanisms to manage the flow of information.

In the readout phase, the final prediction is made by applying the readout function $R$ based on the final hidden states $h_v^T$ of all the vertices according to
\begin{equation}
    \hat{y} = R(\left\{ h_v^T | v\in G\right\}),
\end{equation}
where $R$ consists of a set2set model~\cite{set2set} which generates a graph-level representation $g$ that is invariant to the permutation of input node features. Then, the vector $g$ goes through two fully connected layers to obtain the final prediction $\hat{y}$.

\subsection{Learning objective}\label{sec:LO}\vspace{5pt}
\tcr{In the context of MOCU-based OED, the trained MPNN model serves as a surrogate model that predicts the expected remaining MOCU of each experiment. The predicted MOCU values are then used to prioritize the experiments in the experimental design space based on their expected efficacy and select the optimal experiment as described in~\eqref{eq:OEDProblem}. For this reason, from the perspective of OED, it is necessary for the trained model not only to predict the expected remaining MOCU accurately but also to maintain the relative order of the possible experiments when they are prioritized according to their respective expected remaining MOCU values. In order to enhance the capability of the trained model in preserving this order, we define the overall loss function by combining the mean squared error (MSE) loss $\ell_{MSE}$ with a novel axiomatic constraint loss $\ell_{AC}$ as follows:}
\begin{equation}\label{pl}
    L(\mathcal{A}, y, \hat{y}) = \ell _{MSE} + \lambda \ell _{AC},
\end{equation}
\tcr{where ${\mathcal{A}}$ is the uncertainty class, $y$ is the true MOCU value, $\hat{y}$ is the estimate by the model, and $\lambda \in \left[0, 1 \right]$ is a weight parameter. In practice, $y$ is numerically estimated via the Monte Carlo sampling approach~\cite{Hong2021}.} The MSE loss is defined as a squared difference $\ell_{MSE} = (y - \hat{y})^2$ between the numerical MOCU estimate ${y}$ via the sampling-based approach and the estimate $\hat{y}$ by the proposed model.

In addition to this conventional MSE loss, in this study, we introduce a novel constraint loss that penalizes the deviation from an axiom of MOCU. \tcr{Our axiomatic constraint loss is motivated by the characteristics of MOCU and the MOCU-based OED scheme that are elaborated in Sec.~\ref{MOCU-basedOED}}. As described before, given the outcome of a pairwise synchronization experiment, we either increase the lower bound $a_{i,j}^{\ell}$ or decrease the upper bound $a_{i,j}^u$, as a result of which the objective uncertainty quantified by MOCU is reduced. However, due to the stochastic nature of the numerical estimates obtained by the sampling-based approach~
\cite{Hong2021}, the random fluctuations in the estimated MOCU values may overshadow the actual change in MOCU that results from the updated uncertainty class. We introduce a novel constraint loss to alleviate such fluctuation artifacts. Formally, according to the definition of MOCU in Eq.~(\ref{MOCU}) and the uncertainty class ${\mathcal{A}}$ of the Kuramoto model defined in Sec.~\ref{UK}, the MOCU is an increasing (decreasing) function of the upper bound $a_{i,j}^u$ (lower bound $a_{i,j}^{\ell}$), which can be represented as follows:
\begin{equation}
    \frac{\partial \text{MOCU}}{\partial a_{i,j}^{\ell}} \leq 0 \text{ and } \frac{\partial \text{MOCU}}{\partial a_{i,j}^u} \geq 0  \text{ for }  1\leq i < j \leq N.
\end{equation}
Based on the axiom above, we define the axiomatic constraint loss function as follows:
\begin{equation}
    \ell_{AC} = \sum_{1\leq i < j \leq N}[\max(\frac{\partial \hat{y}}{\partial a_{i,j}^{\ell}}, 0)^2 + \max(-\frac{\partial \hat{y}}{\partial a_{i,j}^u}, 0)^2].
    \label{eq:loss:ac}
\end{equation}
This loss serves as a soft constraint that stabilizes the numerical accuracy of the learned model.

\subsection{Training details}\vspace{5pt}
\tcr{We trained our MPNN based on a two-phase training strategy. First, the MPNN model was trained solely based on a training dataset that contains samples generated by the Kuramoto model with $N=5$ oscillators. In the second phase, we created a training dataset where 50\% of the data points were obtained by randomly sampling the data points from the five-oscillator Kuramoto model dataset and the remaining 50\% was generated based on a seven-oscillator model. This strategy can save the amount of computation needed for creating the training data, as generating the data points for the seven-oscillator Kuramoto model is much more costly compared to generating the data points for the five-oscillator model. The trained MPNN model is then used to perform MOCU evaluation and OED for both uncertain Kuramoto models with five oscillators as well as Kuramoto models with seven oscillators.} In each phase, we trained the model for $400$ epochs \tcr{on the training set} with a batch size of $128$, a learning rate of $0.001$, and Adam optimizer~\cite{DBLP:journals/corr/KingmaB14}. \tcr{The training takes approximately 1 GPU hour for each phase. It should be noted that this does not include the time for generating the training data, which is computationally costly as it involves solving a large number of ordinary differential equations for Kuramoto models with different parameter values.} The strength of the axiomatic constraint loss $\lambda$ was set to $0.0001$ in order to give the model larger freedom at the beginning of the training process and stronger regularization when $\ell_{MSE}$ approaches the same magnitude as $\ell_{AC}$. \tcr{In each epoch, we evaluated the model on a validation set. The best model with the lowest validation loss was then used for performance evaluation and OED simulations in Sec.~\ref{sec:simulation}.}

\section{Simulation Results}\label{sec:simulation}
\tcr{We comprehensively validated our proposed approach based on two different types of Kuramoto models: one type that consists of five oscillators with unknown interaction strength and the other with seven oscillators with unknown interaction strength.
First, we assessed the performance of the proposed MPNN model for predicting the MOCU value for a given uncertainty class and compared it against the latest Monte Carlo sampling approach in a recent study~\cite{Hong2021}.
Second, we validated the efficacy of the proposed axiomatic constraint loss used for training the MPNN model by evaluating its impact on preserving the relative ranking based on the actual MOCU values. Finally, we evaluated the performance of the proposed approach in the context of OED for robust synchronization of uncertain Kuramoto models.
}

\subsection{\tcr{Generation of the training dataset}} \label{data}

\begin{table}[h!]
 \caption{\tcr{Equations and distributions of the parameters used for generating the data points in the training dataset.}}
  \centering
  \begin{tabular}{llll}
    \hline
     & \bfseries Equations &\quad & \bfseries Parameter distribution \\
    \hline
    &$F_{ij} = \frac{|w_i - w_j|}{2}$ &\quad& $w_{i} \sim U[-C, C]$\\
    &$\frac{a_{i,j}^{u}+a_{i,j}^{\ell}}{2} = [b_{ij}d_{i,j}^{strong} + (1-b_{ij})d_{i,j}^{weak}]F_{ij}$ &\quad& $d_{i,j}^{strong} \sim U[0, D_1]$\\
    &$a_{i,j}^{u}-a_{i,j}^{\ell} = 2*d_{i,j}^{uncertain}F_{ij}$ &\quad& $ d_{i,j}^{weak} \sim U[0, D_2]$\\
    & &\quad& $d_{i,j}^{uncertain} \sim U[0, D_3]$\\
    & &\quad& $b_{i, j} \sim \text{Ber}\left(p = 0.5 \right)$\\
    \hline
  \end{tabular}
  \label{Tsetup}
\end{table}

\tcr{Table~\ref{Tsetup} shows the equations and the parameter distributions based on which the training samples were generated.} $U[x,y]$ is the uniform distribution ranging from $x$ to $y$, and ${\text{Ber}\left(p\right)}$ is a Bernoulli random variable determining whether the oscillator pair have strong interaction strength factor $d_{i,j}^{strong}$ or weak interaction strength factor $d_{i,j}^{weak}$. \tcr{For $67\%$ of the training dataset, we set ${b_{i,j}=b_{i,k}} \ ( i\leq j<k\leq N, \forall i$), separating oscillators into two partitions (weak \tcr{or} strong).} For the rest of the dataset, we \tcr{sampled} $b_{i,j}$ independently without such restriction.
We \tcr{generated} a dataset that consists of ${70,000}$ uncertain Kuramoto models with $N=5$ oscillators and ${28,000}$ uncertain Kuramoto models with $N=7$ oscillators. For Kuramoto models with $N= 5$, we use $C=6$, $D_1=1.1$, $D_2=0.6$, and $D_3=0.3$, respectively. For Kuramoto models with $N= 7$, we have $C=10$, $D_1=1.2$, $D_2=0.25$, and $D_3=0.6$, respectively. We \tcr{used} the sampling-based approach~
\cite{Hong2021} with a sample size of $K= 20,480$ \tcr{as a ground truth approach} to compute the MOCU for uncertain Kuramoto models in the training dataset. The resulting dataset \tcr{had} an average MOCU of $0.2378$ (with a standard deviation of $0.21$) for $N=5$, and an average MOCU of $0.7310$ (with a standard deviation of $0.6296$) for $N=7$. The ground truth MOCU \tcr{was} then normalized to zero mean and unit variance \tcr{for training}.
Based on this dataset, ${96 \%}$ of the data points \tcr{were} used for training, and the rest of the data points \tcr{were} used for validation. All experiments \tcr{were} performed on a workstation equipped with \textit{Intel i9-9900K}, 32GB RAM, and \textit{GeForce RTX 2080 Ti}. The Python code of our MPNN model for accelerated MOCU estimation and MOCU-based OED is available at \href{https://github.com/Levishery/AccelerateOED}{https://github.com/Levishery/AccelerateOED}.

\subsection{\tcr{Performance evaluation for MOCU prediction}} \label{acc}
\begin{figure}[h!]
    \centering 
    \includegraphics[width=0.6\textwidth]{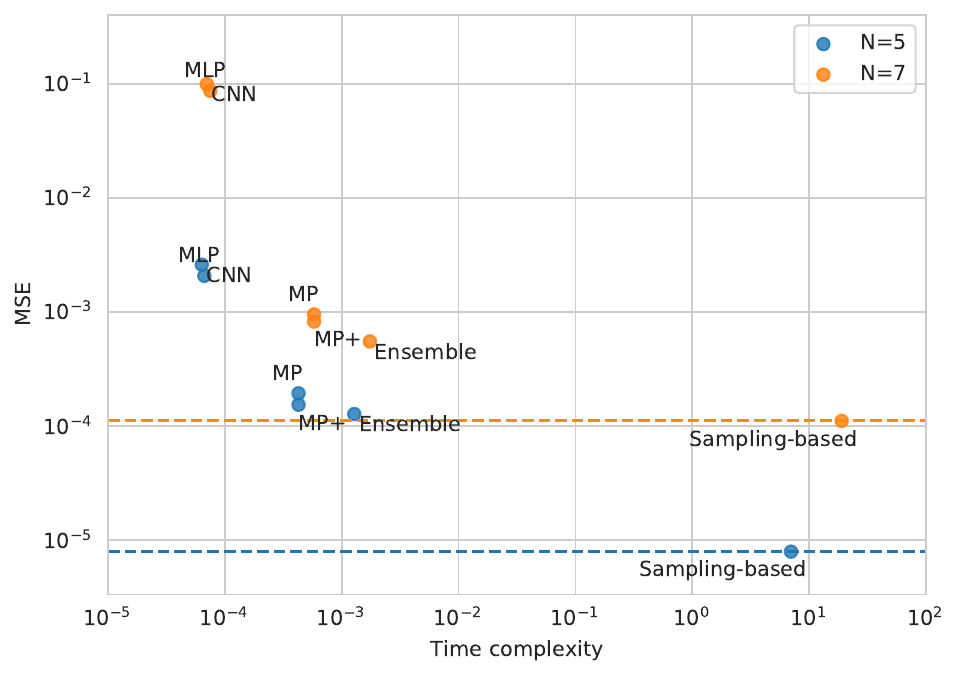}
    \caption{\tcr{Comparison of the time complexity and accuracy among different methods for predicting MOCU.}}
    \label{performance}
\end{figure}

\vspace{5pt}
We \tcr{evaluated} the performance of the proposed model on a test set \tcr{consisting} of $3,000$ uncertain Kuramoto models with $N=5$ oscillators and $2,000$ uncertain Kuramoto models with $N=7$ oscillators, sampled from the same distribution \tcr{described in Table~\ref{Tsetup}}. Figure~\ref{performance} shows the comparison results of the proposed models with benchmark models in terms of time complexity and the MSE on the test dataset. \tcr{MP denotes the MPNN model trained with only the MSE loss, whereas MP+ corresponds to the MPNN model trained with the loss function in \eqref{pl} that incorporates the axiomatic constraint}. \tcr{Ensemble denotes the prediction from an ensemble of three MPNN models, each of which is trained with different random parameter initialization. The ensemble prediction is obtained by taking the average of the predictions made by the different MPNN models in the ensemble.} As a reference, the variance of the sampling-based approach~\cite{Hong2021} on the test set is also shown. For comparison, we trained two other machine learning (ML) models, a convolutional neural network (CNN)~\cite{alexnet} and a multilayer perceptron (MLP). The input of the CNN model is a three-channel $N \times N$ image, representing natural frequency, and interaction strength lower/upper bound, respectively. MLP accepts the input vector of length $N^2$, which is the concatenation of the natural frequency, the upper triangle of the lower/upper bound matrix. Note that, unlike the input for the MPNN, the dimension of the input for the CNN and MLP is fixed, which requires training separate ML models for the MOCU estimation of uncertain Kuramoto models with different numbers of oscillators.

As shown in Fig.~\ref{performance}, the proposed models (MP, MP+, and Ensemble MP+) \tcr{outperformed} the baseline neural networks (i.e., CNN and MLP) by significant margins in terms of prediction accuracy. This \tcr{was} likely due to the effective inductive bias~\cite{47094} introduced by the proposed models. In fact, the inductive bias of the MPNN limits the search space of the models to be invariant to node and edge permutations, which is beneficial for graphical data. During training, we \tcr{observed} that all the deep learning models converge to similar training MSE loss, while the test accuracy significantly differs across different models, as shown in Fig.~\ref{performance}. The results in Fig.~\ref{performance} clearly indicate the generalizability of the MPNN. Our proposed message-passing models tended to yield accurate predictions that were very close to those obtained by the original sampling-based approach. However, the proposed models were significantly faster than the sampling-based approach, resulting in four to five orders of magnitude speed-up. Especially, the speed gains were more prominent for the larger system. For example, for Kuramoto models with ${N=7}$ oscillators, the sampling-based approach took $19$ seconds on average to estimate the MOCU for a given uncertain Kuramoto model. On the contrary, it took only an average of $1.160$ seconds for the MPNN model to estimate MOCU for the entire test set ($2,000$ data points). This amounts to a $32,758$ speed-up. We expect that the relative computational efficiency of the proposed method with respect to the sampling-based approach will further increase for more complex Kuramoto models with a larger number of oscillators. Additionally, results shown in Fig.~\ref{performance} show that the constraint loss $\ell_{AC}$ improves the prediction accuracy. \tcr{It is worth noting that the main advantage of incorporating this loss function into the learning objective is its ability to maintain the \tcr{relative rank of the MOCU} of each experiment, which will be discussed further in Sec.~\ref{sec:effect}.} The results also showed that the ensemble scheme is beneficial for enhancing accuracy at an increased computational cost.

\tcr{Besides, we conducted additional experiments to further compare the performance of the trained MPNN model against other popular models designed for graph-based regression tasks~\cite{corso2020principal,Ying2021DoTR}. Specifically, we trained and tested Principal Neighbourhood Aggregation (PNA)~\cite{corso2020principal} that replaces the sum aggregator in the MPNN model with a degree-scaled aggregator and Graphormer~\cite{Ying2021DoTR} that incorporates structural encoding schemes (such as centrality encoding and spatial encoding) that leverage structural information of a graph reflected on nodes and their neighbors. For Graphormer, we tested two representative architectures $Graphormer_{SLIM}$ and $Graphormer_{BASE}$. The simulation results indicated that there is no noteworthy difference among the considered models in predicting MOCU. We also investigated the efficacy of the proposed axiomatic constraint loss by incorporating it into the training of the PNA model along with the MSE loss. As expected, the proposed axiomatic constraint loss was shown to be effective in improving the predictive performance of the PNA model. Detailed performance comparison results can be found in the supplementary information.}

\subsection{Effect of axiomatic constraint loss on regression accuracy}
\label{sec:effect}\vspace{5pt}
\tcr{As discussed in Section~\ref{sec:LO}, for the surrogate models to be effective in the context of MOCU-based OED, they need to be capable of maintaining the order of experiments in terms of their true expected remaining MOCU. If the (expected remaining) MOCU predicted by the surrogate models significantly changes the priority of the available experiments--compared to the case when the experiments are prioritized based on the true (expected remaining) MOCU--adopting such surrogate models would result in the degradation of the efficacy of the selected experiments.
In the MOCU-based OED loop, the uncertainty class is updated based on the experimental outcome of the selected experiment. In our application, we reduce the range of uncertain interaction strength $a_{i,j} \in [a_{i,j}^{\ell},a_{i,j}^u]$ based on the experimental outcome (i.e., whether the selected oscillator pair is synchronized or not) of the selected oscillator pair $(i,j)$: either by increasing the lower bound $a_{i,j}^{\ell}$ if $(i,j)$ gets synchronized or by decreasing the upper bound $a_{i,j}^u$, otherwise. While the extent of uncertainty reduction depends on various factors, MOCU should be a non-increasing function of model uncertainty. In other words, when the uncertain range decreases for one of the interaction strengths $a_{i,j}$, MOCU should decrease or stay the same (if the reduced range for $a_{i,j}$ does not affect the task--i.e., robust synchronization), but never increase. In that regard, we assessed the performance of the proposed MPNN model trained using the loss function in \eqref{pl} in terms of its ability to preserve the order of MOCU values as we change $a_{i,j}^{\ell}$ or $a_{i,j}^u$.} 

\begin{table}[h]
\renewcommand{\arraystretch}{1.5}
 \caption{\tcr{Evaluation of different methods for their respective ability to preserve ranking.}}
  \centering
  \begin{tabular}{ccccc}
    \hline
    \bfseries  Number of oscillators & \multicolumn{2}{c}{\bfseries $N = 5$} & \multicolumn{2}{c}{\bfseries $N = 7$}\\
    \hline
    \bfseries Method   & \bfseries $ a_{i,j}^{\ell} \uparrow$  & \bfseries $a_{i,j}^u \downarrow$   &\bfseries $a_{i,j}^{\ell} \uparrow$    & \bfseries $a_{i,j}^u\downarrow$\\
    \hline
    Sampling-based & $0.6599$   & $0.6543$   & $0.5720$   & $0.5660$\\
    MP & $0.8553$   & $0.8466$  &  $0.7540$  & $0.7600$   \\
    MP+ & $0.9254$ & $0.9178$   &  $0.8660$ & $0.8475$   \\
    \hline
  \end{tabular}
  \label{rank}
\end{table}

Table.~\ref{rank} summarizes the performance of the different MOCU estimation schemes in terms of their ability to preserve the ordering of the MOCU values--and ultimately, the ranking of the potential experiments according to their expected efficacy in reducing MOCU. We used the same test data described in~\ref{acc} except for reducing the dataset size to one-fifth for the sampling-based approach due to its high computational cost. Analogous to updates in OED, for each uncertain Kuramoto model in the test set, we randomly selected oscillator pair $(i,j)$ and then increased $a_{i,j}^{\ell}$ or decreased $a_{i,j}^u$ to $(1/2)*(a_{i,j}^{\ell}+a_{i,j}^u)$, which are denoted as $a_{i,j}^{\ell} \uparrow$ and $a_{i,j}^u \downarrow$ in Table.~\ref{rank}, respectively. As the performance metric, we measured the percentage of the test data for which a given model yields lower MOCU for the reduced uncertainty class as expected.

As we can see in Table~\ref{rank}, the MPNN models (both MP and MP+) were very effective in preserving the ordering of the estimated MOCU values for different lower and upper bounds. This is likely due to the stochastic nature of the numerical estimates obtained by the sampling-based approach, where the statistical deviation between the estimates may be relatively larger compared to the actual difference between the MOCU values for small changes in the bounds $a_{i,j}^{\ell}$ and $a_{i,j}^u$. The results in Table~\ref{rank} show that the message-passing network model tends to effectively regularize the statistical fluctuations in the MOCU estimates in the training data generated by the sampling-based scheme~\cite{Hong2021}. Furthermore, as we would expect, incorporating the axiomatic constraint loss $\ell_{AC}$ in Eq.~(\ref{eq:loss:ac}) into the learning objective clearly improves the preservation of the ordering of the MOCU estimates. Our results demonstrate the solid ability of our proposed MPNN model to preserve the MOCU-based ordering of different uncertainty classes. This is a critical property as a surrogate model for MOCU-based experimental design since it allows us to reliably prioritize potential experiments in the experimental design space by accurately quantifying the expected impact of a given experiment on reducing model uncertainty.

\subsection{Evaluation of OED performance}\label{OEDsimu}\vspace{5pt}

\tcr{Finally, we verified the efficacy of the proposed approach by applying the MPNN model to the OED loop. Our goal was to assess how effective the experiments selected using the MPNN-based scheme were in reducing the model uncertainty and how its performance compares to other OED schemes.} For simulations, we used the uncertainty classes considered in~\cite{Woo2021}. Specifically, we assumed that the $N=5$ oscillator model has natural frequencies $\boldsymbol{\omega} = [-2.50, -0.6667, 1.1667, 2.0, 5.8333]^T$. The uncertainty class is defined as:
\begin{equation}\label{U5}
\mathbf{a}^u =
\small{
\!\begin{aligned}
&
\left[\begin{matrix}
  1.0541 & 0.6325 & 0.7762 & 1.4375 & 1.0542 & 0.6900 & 1.6819 & 0.4791 & 2.6833 & 2.2041
\end{matrix}\right]^T,
\end{aligned}
}
\end{equation}
\begin{equation}\label{L5}
\mathbf{a}^{\ell} =
\small{
\!\begin{aligned}
&
\left[\begin{matrix}
  0.7791 & 0.4675 & 0.5737 & 1.0625 & 0.7792 & 0.5100 & 1.2431 & 0.3541 & 1.9833 & 1.6291 
\end{matrix}\right]^T.
\end{aligned}
}
\end{equation} 
For the $N=7$ oscillator model, we assumed that the natural frequencies of the oscillators were defined as $\boldsymbol{\omega} = [-3.46, -1.96, -0.68, -0.38, -0.37, 6.12, 8.3287]^T$. The uncertainty class was defined as:
\begin{equation}\label{U7}
\mathbf{a}^u =
\small{
\!\begin{aligned}
&
\left[\begin{matrix}
  0.848 & 0.988 & 1.446 & 1.607 & 3.820  & 0.915 & 0.400 &
  0.850 & 0.419 & 4.162
\end{matrix}\right.\\[-4pt]
&\qquad\qquad
\left.\begin{matrix}
   1.090 & 0.122 & 0.039 & 2.124 & 0.872 & 0.007 & 2.737 & 1.804 & 1.360  & 0.744 \quad 1.174
\end{matrix}\right]^T,
\end{aligned}
}
\end{equation}
\begin{equation}\label{L7}
\mathbf{a}^{\ell} =
\small{
\!\begin{aligned}
&
\left[\begin{matrix}
  0.073 & 0.172 & 0.153 & 0.054 & 0.501 & 0.463 & 0.043 & 0.015 & 0.096 & 0.501
\end{matrix}\right.\\[-4pt]
&\qquad\qquad
\left.\begin{matrix}
  0.103 & 0.007 & 0.009 & 0.139 & 0.408 & 0.000 & 0.131 & 0.119 & 0.300 & 0.286 \quad 0.131
\end{matrix}\right]^T.
\end{aligned}
}
\end{equation}
Note that these data points are covered by the distribution of the training data used to train our MPNN models. The natural frequency of the external control oscillator is set to the average frequency of the oscillators in the model.

\begin{figure}[!t]
\centering
\subfigure[Kuramoto model with 5 oscillators]{
\includegraphics[width=0.48\textwidth]{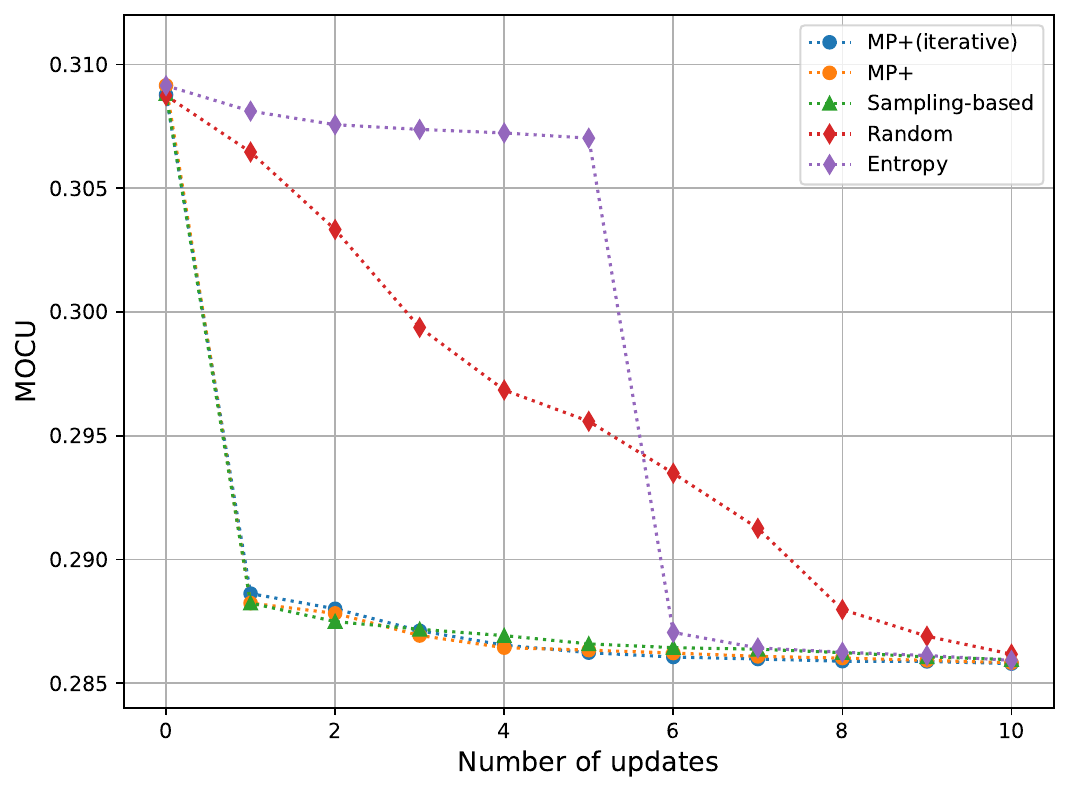}
}
\subfigure[Kuramoto model with 7 oscillators]{
\includegraphics[width=0.48\textwidth]{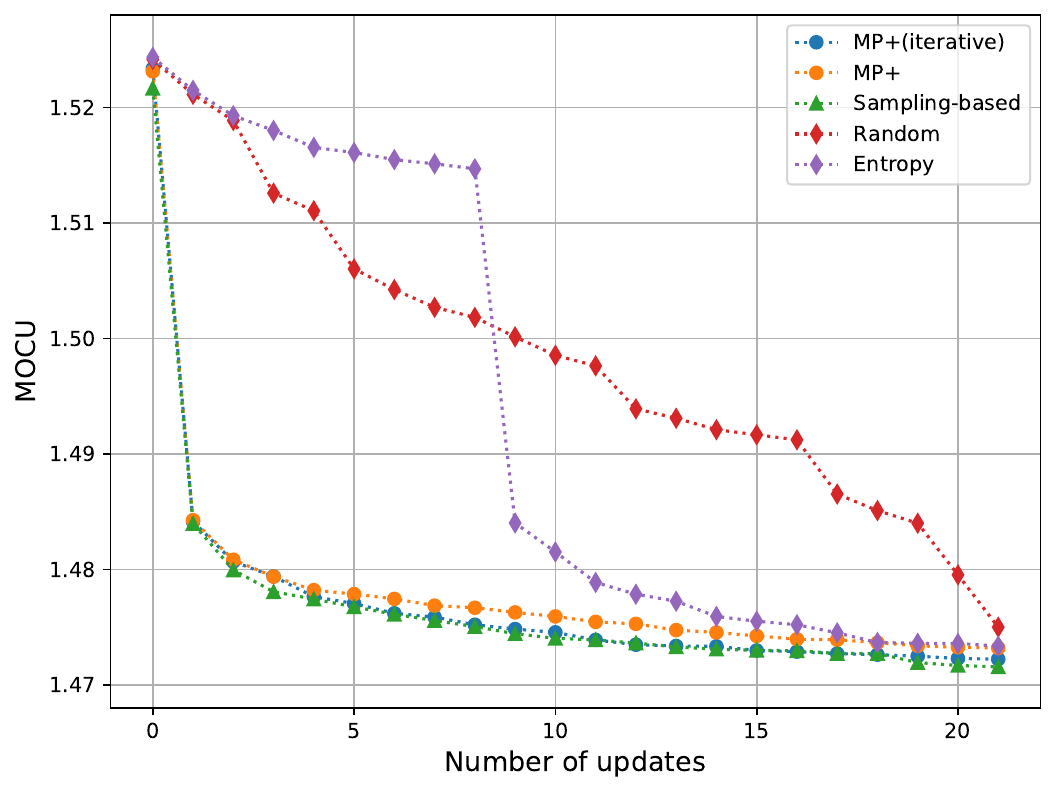}
}
\caption{\tcr{Objective model uncertainty (quantified by MOCU) decreases with sequential experimental updates. Various experimental design strategies are compared to assess their respective efficacy in prioritizing advantageous experiments in the design space.}}
\label{simu5}
\end{figure}


For each uncertainty class, we determined a series of experiments based on a specific experimental design scheme and assessed how each experimental update reduces model uncertainty quantified by MOCU. For comparison, we evaluated the performance of the proposed MPNN-based scheme, the original sampling-based scheme in~\cite{Hong2021}, the entropy-based scheme, and the random selection approach. In the MPNN-based scheme, we estimated the expected remaining MOCU for all $N \choose 2$ experiments in the experimental design space using the MPNN model, which was then used to prioritize the experiments. Starting from the experiment with the smallest expected remaining MOCU, we performed the selected experiment, updated the uncertainty class based on the outcome, and continued this process with the next best experiment until all experiments in the experimental space were exhausted. In the sampling-based scheme, MOCU was numerically estimated via sampling as described in~\cite{Hong2021}, where the experimental sequence was determined in a similar fashion based on the remaining expected MOCU. We also considered an iterative scheme based on our MPNN model, where the expected remaining MOCU values of the available experiments were re-estimated after each experimental update, based on which the next experiment was selected in an iterative manner. Due to the formidably high computational cost, the sampling-based scheme was not considered for iterative experimental design. In the entropy-based approach, we prioritized the experiments based on the uncertain range $a_{i,j}^u-a_{i,j}^{\ell}$, from the experiment for the pair with the largest uncertain range to the smallest. Finally, in the random selection approach, we performed experimental updates in a randomized order. \tcr{For reliable estimation of the MOCU at each experimental update, we numerically computed the MOCU values $10$ times using the sampling-based scheme and used the average. In each MOCU computation, a sample size of $K=20,480$ was used. The overall experiments were repeated $70$ times by randomly sampling the true (unknown) Kuramoto model from the uncertainty class.}

\tcr{Figure~\ref{simu5} shows the performance evaluation results of the various experimental design schemes. For each scheme, the corresponding curve shows how the model uncertainty (quantified by MOCU) decreases as the number of experimental updates increases. Results are shown for uncertain Kuramoto models that consist of ${5}$ oscillators (see Fig.~\ref{simu5}(a)) and uncertain Kuramoto models with ${7}$ oscillators (see Fig.~\ref{simu5}(b)). As shown in Fig.~\ref{simu5}, the sampling-based approach achieved the best OED performance in reducing the model uncertainty. The proposed MPNN-based scheme (MP+) yielded virtually identical OED performance compared to the sampling-based scheme in both simulations. However, the computational gain of adopting the MPNN model for MOCU-based OED is huge, as it can accelerate the computational speed of OED by four to five orders of magnitude for the Kuramoto models considered in this study. For example, for $N=7$, it took an average of $2,995.8$ seconds (with a standard deviation of $0.4988$) for the original sampling-based OED scheme to compute the expected remaining MOCU values for all possible experiments for a single experimental update. In comparison, it took only an average of $0.0340$ seconds (with a standard deviation of $0.0030$) for our proposed MPNN model. Similarly, for $N=5$, the computation of the expected remaining MOCU values for all possible experiments for a single experimental update took an average of $1,345.4$ seconds (with a standard deviation of $0.5262$) when the sampling-based OED scheme was used. On the other hand, the MP model needed only an average of $0.0321$ seconds (with a standard deviation of $0.0019$) for the same task.
The random and entropy-based experimental design approaches, in contrast, failed to effectively reduce the model uncertainty within the first few experiments. On average, the random selection scheme resulted in a linear decrease of the model uncertainty with respect to the number of updates, as one would expect. The entropy-based scheme, which prioritizes the experiment that can reduce the uncertainty of the interaction strength $a_{i,j}$ with the largest uncertain range $a_{i,j}^u-a_{i,j}^\ell$, was not able to initially reduce the model uncertainty by a significant amount even after several experimental updates. This clearly illustrates why it is important for OED techniques to focus on reducing the model uncertainty that practically matters, which can only be done by quantifying how the uncertainty impacts the operational objective (e.g., achieving robust global frequency synchronization of an uncertain Kuramoto model with minimum control cost).
} 

Interestingly, the iterative MPNN-based OED scheme did not result in performance improvement in our simulations. The potential underlying reason for this observation may be found in the MOCU reduction trends shown in Fig.~\ref{simu5}. For both $N=5$ and $N=7$, a single experiment--if properly selected by a MOCU-based OED scheme--is able to effectively reduce the objective model uncertainty. As a result, the subsequent experiments have only a relatively small impact on uncertainty reduction. However, we expect that an iterative OED scheme would be beneficial in more complicated scenarios, where multiple experiments would be needed to effectively reduce the overall model uncertainty.

\section{Concluding Remarks}\label{sec:conclusion}

In this work, we proposed a novel approach for accelerating MOCU-based objective-UQ and OED based on neural message-passing. To the best of our knowledge, this is the first study that demonstrates the possibility of designing accurate data-driven surrogate models for efficient objective-UQ and OED based on the concept of MOCU. As demonstrated, based on uncertain Kuramoto models that consist of interacting oscillators whose coupling strengths are not known with certainty, the proposed approach significantly accelerates MOCU estimation and MOCU-based OED. Compared to the original sampling-based approach~\cite{Hong2021}, the proposed model was shown to improve the speed by four to five orders of magnitude for Kuramoto models with $N=5$ or $N=7$ oscillators. We expect that the overall gain in terms of computational efficiency will be even higher for a more complex Kuramoto model with a larger number of oscillators. 


It is worth noting that training the neural network requires generating sufficient training data, where the original sampling-based approach~\cite{Hong2021} is used for MOCU computation for various uncertain Kuramoto models. However, once the training data are generated, actual training of the MPNN requires a relatively short time and the trained MPNN can be used for extremely fast and accurate MOCU estimation for various uncertain Kuramoto models. Considering that the OED process typically requires a large number of MOCU computations, our proposed scheme has the added benefit of moving the heavy computational cost of MOCU evaluations from online to offline. Nevertheless, it would be an interesting research direction to apply the proposed axiomatic constraint loss in a \textit{self-supervised} manner to reduce the model's dependency on the amount of labeled training data. Similar to the idea of constructive learning in~\cite{chen2020simple}, we may pre-train the deep learning model by randomly generating paired uncertainty classes and encouraging the model to predict the correct ranking, without using any ground truth labels. We expect that such a pre-training procedure may provide a good starting point for the model that can ultimately reduce the amount of labeled data needed to train a high-quality surrogate model.


In this study, we have mainly focused on applying the proposed MPNN to the objective-UQ and OED for uncertain Kuramoto models in order to demonstrate its advantages over the state-of-the-art, recently presented in~\cite{Hong2021}. However, it should be noted that the proposed model architecture and the learning scheme can be easily adapted to accelerate MOCU estimation and MOCU-based OED for different types of \textit{systems} (i.e., beyond Kuramoto oscillator models) and other operational \textit{objectives} (i.e., beyond robust synchronization or control), and a wide range of \textit{experimental design} problems. \tcr{For example, it would be interesting to apply the proposed MPNN-based approach to accelerate MOCU-based Bayesian active learning (BAL). As proposed in~\cite{Zhao2021ICLR,Zhao2021AISTATS}, soft-MOCU (SMOCU) or weighted-MOCU (WMOCU)--which are variants of MOCU--can be used to guide the pool-based active learning process in order to increase the labeling efficiency and to effectively improve the accuracy of optimal Bayesian classifiers with a relatively small number of label acquisitions. By incorporating a DL-based surrogate model for MOCU estimation (or the evaluation of SMOCU/WMOCU-based acquisition functions), the overall computational efficiency of the BAL process could be significantly accelerated without degrading the learning outcomes.}

Considering the enormous OED speed improvement that can be attained by our proposed MPNN-based method, we can envision the application of MOCU-based OED for experimental design in real-world complex uncertain systems, which was previously intractable due to the high computational cost. For example, Kuramoto oscillators are known to be able to model the brain network synchronization phenomena~\cite{Mohseni2017,Gilmer2017,kitzbichler2009broadband,ferrari2015phase,WEINRICH20173061,10.1371/journal.pcbi.1006978}. Many cognitive and behavioral states, including perception, memory, and action, have been described as the emergent properties of coherent or phase-locked oscillations in transient neuronal ensembles. As shown in~\cite{WEINRICH20173061}, non-invasive transcranial alternating-current stimulation(tACS) at the beta frequency ($20$ Hz) can increase local activity in the beta band and modulate the connectivity pattern of the stimulated motor cortex. In the aforementioned study~\cite{WEINRICH20173061}, they model tACS as a controlling oscillator in the Kuramoto model similar to our work. The model was able to faithfully reproduce key features of the experimental data. Recently, the Kuramoto model was used to study the relationship between a brain network's global synchronization and strong connections over a long distance, where the connectivity matrix was constructed from viral tracing experiments~\cite{10.1371/journal.pcbi.1006978}. This research area provides promising venues for applying our MPNN-based scheme proposed in this work. It is our plan to explore the potential application of the accelerated objective-UQ and OED scheme via message-passing networks to the optimal therapeutic stimulation of brain networks under uncertainty.

\section*{Acknowledgment}

The work of H.-M. Woo and B.-J. Yoon was supported in part by the NSF Award 1835690 and the DOE Award DE-SC0019303.

\bibliographystyle{unsrt}  
\bibliography{Reference}  

\clearpage
\begin{flushleft}
{\Large
\textbf{Supplemental Material for ``Neural message passing for objective-based uncertainty quantification and optimal experimental design''}
}
\end{flushleft}
\section*{Performance comparison of various deep learning models for MOCU prediction}
\begin{table}[h]
\renewcommand{\arraystretch}{1.5}
 \caption{Performance comparison of various deep learning models for predicting MOCU.}
  \centering
  \begin{tabular}{c|c|cc}
    \hline
    \bfseries    & \#params & \multicolumn{2}{c}{test MSE$(*10^{-4})$}\\
    \hline
    \bfseries Methods  &  & N=5 & N=7\\
    \hline
    MP~\cite{Gilmer2017} & $154,593$ & $1.930$   & $9.483$
    \\
    MP+ & $154,593$ & $\boldsymbol{1.534}$   & $8.226$  \\
    \hline
    PNA~\cite{corso2020principal} & $675,902$ &  $2.226$ & $9.836$ \\
    PNA+ & $675,902$ & $1.925$ & $\boldsymbol{7.966}$ \\
    \hline
    $\text{Graphormer}_\text{{SLIM}}$(batchsize$=128$)~\cite{Ying2021DoTR} & $578,241$ & $5.196$ & $15.53$ \\
    $\text{Graphormer}_\text{{SLIM}}$(batchsize$=1024$)~\cite{Ying2021DoTR} & $578,241$ & $1.663$ & $9.640$ \\
    \hline
  \end{tabular}
  \label{backbone}
\end{table}

\end{document}